\documentclass[letterpaper, 10 pt, conference]{ieeeconf}  

\IEEEoverridecommandlockouts                              

\usepackage{soul}
\usepackage[utf8]{inputenc} 
\usepackage[T1]{fontenc}    
\usepackage{hyperref}       
\usepackage{url}            
\usepackage{booktabs}       
\usepackage{amsfonts}       
\usepackage{nicefrac}       
\usepackage{microtype}      
\usepackage{xcolor}         
\newcommand{\CommentColor}[1]{\textcolor{blue!70!black}{// #1}}
\usepackage{graphicx}
\usepackage{float}
\usepackage{amsmath}
\usepackage{caption}
\usepackage{subcaption}
\usepackage{listings}
\usepackage{mathtools}
\usepackage{lineno}
\usepackage{mathrsfs}
\usepackage{wrapfig} 
\usepackage{algorithm}
\usepackage{tabularx}
\usepackage[table]{xcolor}
\usepackage{booktabs}
\usepackage{multirow}
\usepackage{algorithmic}
\renewcommand{\mathbb}[1]{\mathbf{#1}}
\newcommand{\ind}{\mathbb{1}}
\usepackage{color}
\definecolor{polina_purple}{rgb}{0, 0, 0}
\newcommand{\polina}[1]{\textcolor{polina_purple}{#1}}

\newcommand{\edit}[1]{{\color{black} #1}}

\title{\LARGE \bf
Towards Automated Semantic Interpretability in Reinforcement Learning via Vision-Language Models
}

\author{Zhaoxin Li\textsuperscript{* 1} \quad
Zhang Xi-Jia\textsuperscript{* 1} \quad
Batuhan Altundas\textsuperscript{1} \quad
Letian Chen\textsuperscript{1} \quad 
Rohan Paleja\textsuperscript{2} \quad
Matthew Gombolay\textsuperscript{1} 
\thanks{1. Georgia Institute of Technology, Atlanta, GA 30332, USA}%
\thanks{2. Purdue University, West Lafayette, MA 47907}%
}

\begin{document}

\maketitle
\thispagestyle{empty}
\pagestyle{empty}

\begin{abstract}
Semantic interpretability in Reinforcement Learning (RL) enables transparency and verifiability of decision-making. Achieving semantic interpretability in reinforcement learning requires (1) a feature space composed of human-understandable concepts and (2) a policy that is interpretable and verifiable. However, constructing such a feature space has traditionally relied on manual human specification, which often fails to generalize to unseen environments. Moreover, even when interpretable features are available, most reinforcement learning algorithms employ black-box models as policies, thereby hindering transparency. We introduce \textbf{i}nterpretable \textbf{T}ree-based \textbf{R}einforcement learning via \textbf{A}utomated \textbf{C}oncept \textbf{E}xtraction (iTRACE), an automated framework that leverages pre-trained vision-language models (VLM) for semantic feature extraction and train a interpretable tree-based model via RL. To address the impracticality of running VLMs in RL loops, we distill their outputs into a lightweight model. By leveraging Vision-Language Models (VLMs) to automate tree-based reinforcement learning, iTRACE loosens the reliance the need for human annotation that is traditionally required by interpretable models. In addition, it addresses key limitations of VLMs alone, such as their lack of grounding in action spaces and their inability to directly optimize policies. We evaluate iTRACE across three domains: Atari games, grid-world navigation, and driving. The results show that iTRACE outperforms other interpretable policy baselines and matches the performance of black-box policies on the same interpretable feature space.

\end{abstract}

\vspace{-0.4em}
\section{Introduction}
\vspace{-0.3em}

Reinforcement learning (RL) agents operate in dynamic environments, often developing policies that rely on latent representations not directly interpretable by humans~\cite{Singh2022survey}. Ensuring the learned representations align with human-understandable concepts, a goal often referred to as semantic interpretability, is crucial for ensuring safety and preventing unintended harmful actions. While progress has been made in interpretable AI, achieving semantic interpretability in RL remains particularly challenging~\cite{rudin2021interpretablemachinelearningfundamental}. \edit{It requires (1) constructing a feature space composed of human-understandable concepts, a process that has traditionally relied on human annotation and lacks generalizability. (2) In addition, it requires learning policies that are both high-performing and transparent~\cite{Glanois2024survey}.}


To learn an interpretable model on a semantically interpretable feature space, various model classes can be considered, including decision trees and rule-based policies, both of which facilitate explicit reasoning about an agent's actions~\cite{Glanois2024survey}. Decision trees have been used to represent policies in a structured, human-readable format~\cite{silva2021encoding, paleja2023interpretable}. 
Rule-based methods offer an alternative by encoding policies in a form that can be explicitly followed and verified~\cite{delfosse2024interpretable, jiang2019neural}. 
However, none of these prior works address high-dimensional image or video inputs; instead, they rely either on human-defined semantic concepts or ground-truth feature vectors provided by the environment.

\begin{figure}[t]
  \centering
  \caption{High-level overview of iTRACE.}
  \includegraphics[width=\columnwidth]{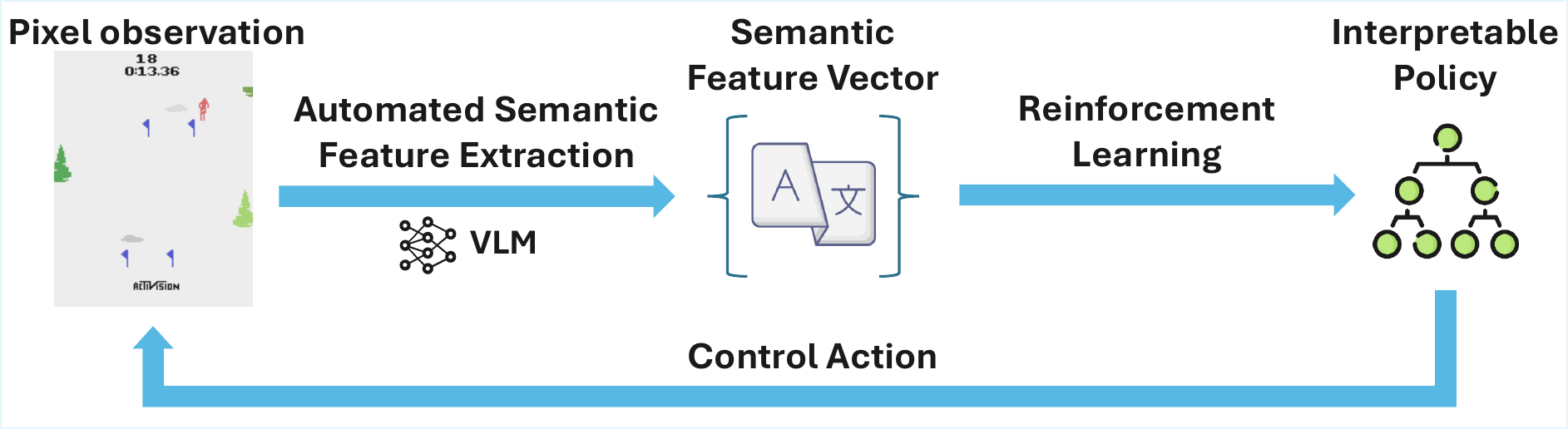}
  
  \vspace{-2em}
  \label{fig:myimage}
\end{figure}


\edit{In this paper, we tackle the challenge of mapping high-dimensional observations into human-understandable features that can serve as a low-dimensional input space for interpretable decision-making models. Prior work has attempted to bridge the gap between raw image inputs and interpretable features~\cite{arulkumaran2017deep}, but existing methods fall short of true semantic interpretability. Some approaches generate mathematical or visual representations that do not correspond to human-understandable concepts~\cite{Liu2015feature,greydanus2018visualizing}. Others rely on predefined feature spaces that fail to generalize across environments~\cite{doumanoglou2023unsupervisedinterpretablebasisextraction, scabini2024comparativesurveyvisiontransformers}.
As a result, human involvement remains central: researchers either manually define semantic features~\cite{Zytek2022need} or label machine-generated ones~\cite{templeton2024scaling}. However, this reliance on human annotation is not automated and incurs significant time costs, limiting scalability to new domains.}

The emergence of Vision-Language Models (VLMs) has opened new possibilities for addressing these challenges, particularly the difficulty of building human-aligned feature extractor. VLMs, known for their capabilities in visual understanding, language reasoning, and decision-making ~\cite{yang2023foundationmodelsdecisionmaking}, have seen widely used in various tasks ~\cite{yang2023foundationmodelsdecisionmaking,zhai2024finetuninglargevisionlanguagemodels}. 


In this work, we propose \textbf{i}nterpretable \textbf{T}ree-based \textbf{R}einforcement learning via \textbf{A}utomated \textbf{C}oncept \textbf{E}xtraction (iTRACE). iTRACE leverages VLMs to extract semantic features from raw pixel inputs and trains Interpretable Control Trees to provide the rationale behind RL decisions in terms of human-understandable concepts. Our framework (i) achieves performance competitive with black-box models while surpassing uninterpretable models built on the same feature space, (ii) supports both \edit{single-frame and multi-frame inputs, enabling the capture of temporal information}, (iii) generalizes to unseen environments without human involvement, and (iv) ensures computational efficiency by training a lightweight vision model as an alternative to VLM-based feature extraction. To the best of our knowledge, iTRACE is the \textbf{first automated framework} to achieve semantic interpretability in RL while generalizing effectively to new environments without human-in-the-loop.


\section{Related work}

\subsection{Interpretable reinforcement learning}

\begin{figure*}[h!]
    \centering
    \includegraphics[width=0.9\textwidth]{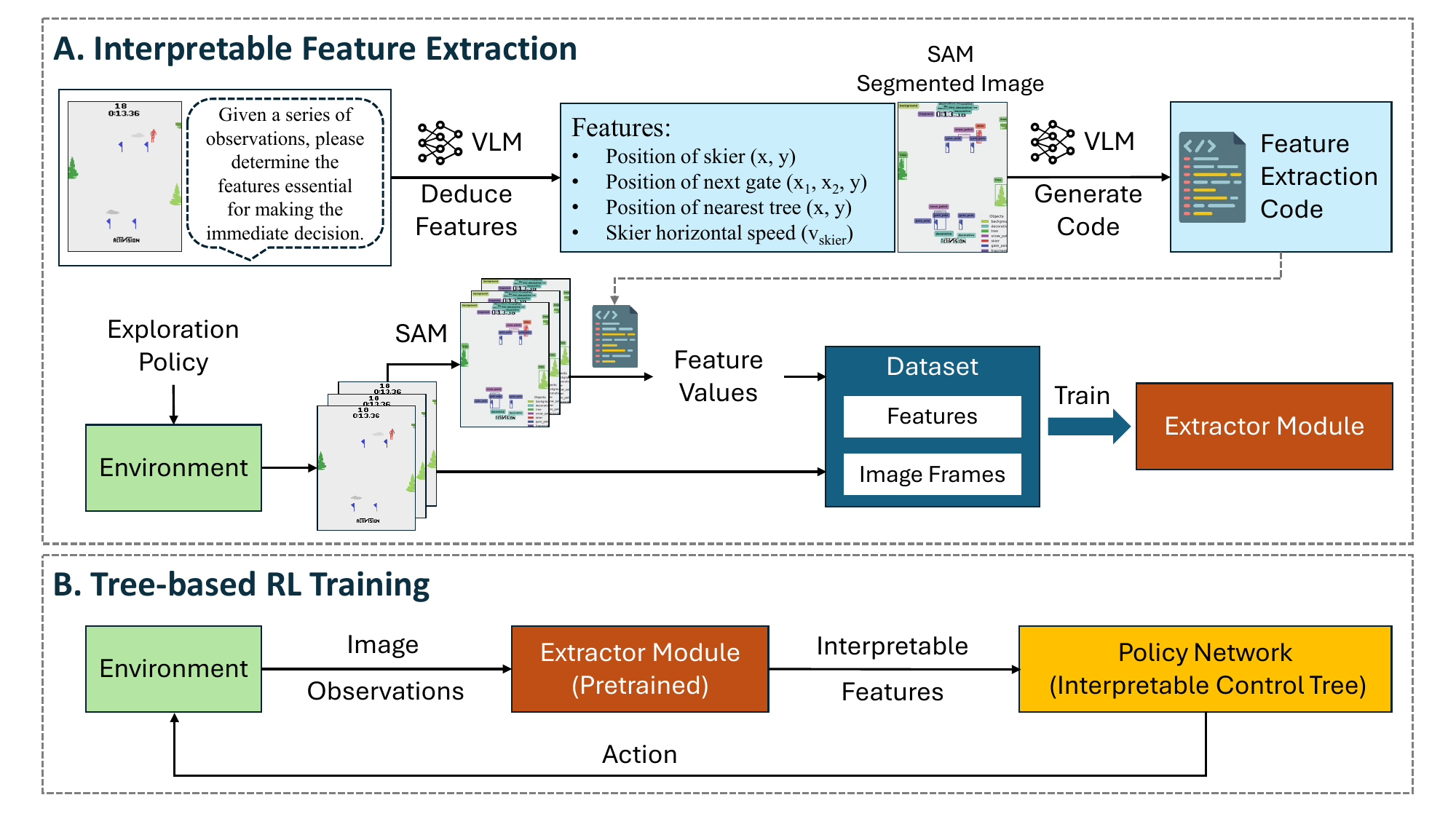}
        \caption{Overview of \textbf{iTRACE}, our proposed framework. \textbf{A. Feature Extraction} follows three steps: (1) the VLM identifies key semantic features from domain knowledge and observations; (2) it generates code to extract features from SAM-based segmentations, extending to multi-frame inputs if needed; and (3) an exploration policy generates feature–image pairs, which are used to train a lightweight Extractor Module to replace costly SAM queries during RL. \textbf{B. Tree-based RL Training} then leverages the Extractor Module to obtain interpretable features, which serve as inputs to an Interpretable Control Tree for policy learning.  }
    \label{fig:framework_overview}
\end{figure*}

\polina{
Interpretable RL remains an open challenge, as current methods struggle to combine human-understandable policies with high performance, to determine when such policies are feasible, and to maintain transparency in high-dimensional state spaces~\cite{rudin2021interpretablemachinelearningfundamental}. 
Following Rudin’s definition, we view Interpretable RL as learning constrained models whose reasoning humans can directly understand, distinct from post-hoc explainability tools such as SHAP~\cite{beechey2023explaining_rl_shap}, LIME~\cite{lu2025explainable}, or saliency maps~\cite{shi2022selfsupervised,wang2020attribution}, which are often misleading and susceptible to hallucinations. 
}

\polina{Our work addresses the limitation of prior tree-based methods, which previously required manual feature engineering or were limited to structured inputs. Among inherently interpretable models, tree-based methods stand out for producing explicit human-readable rules~\cite{quinlan1987generating}, handling categorical and nonlinear interactions~\cite{loh2011classification}, supporting sparsity constraints~\cite{hu2019optimal}, and achieving accuracy competitive with black-box models when properly optimized~\cite{grinsztajn2022tree,mcelfresh2023neural,zhou2019deep}. 
Recent advances extend these strengths: Differentiable Decision Trees (DDTs) enable gradient-based optimization of interpretable models~\cite{silva_optimization_2020}; Interpretable Continuous Control Trees (ICCTs) extend this approach to continuous control tasks, balancing transparency and performance~\cite{paleja_learning_2023}; and POETREE introduces dynamically growing probabilistic trees that evolve in complexity during training~\cite{pace2022poetree}.
}

\subsection{Feature extraction for reinforcement learning}

Nonparametric methods (e.g., manifold and spectral learning) and nonlinear parametric methods (e.g., deep learning) are commonly used in RL for feature extraction but typically do not produce interpretable features~\cite{Liu2015feature}.
\polina{
Other approaches focuses on identifying visual representations that boost performance without guaranteeing human-meaningful concepts: SSINet highlights task-relevant regions with attention masks~\cite{shi2022selfsupervised}, while perturbation-based saliency maps identify influential pixels~\cite{greydanus2018visualizing}.
Yet, few works construct semantically meaningful feature spaces without human supervision for RL training. For example, Wu et al.~\cite{wu2024readreaprewardslearning} automatically parse features from Atari manuals, but their extracted features remain noisy and still rely on oracle human-defined information. In contrast, our framework automatically defines and extracts features without human-in-the-loop assumptions.
}

\section{\edit{Interpretable Feature Extraction}}


\edit{In this section, we introduce the interpretable feature extraction component of the iTRACE framework, shown in the first block of Fig.~\ref{fig:framework_overview}.} This component leverages the visual comprehension and reasoning capabilities of VLMs to automate feature extraction through a three-step pipeline: (1) \textbf{zero-shot} prompting is used to identify semantic features that deterministically influence decision-making \edit{(line 3 in Algorithm~\ref{alg:itrace})}; (2) feature extraction code is generated to capture both static and temporal information \edit{(line 4 in Algorithm~\ref{alg:itrace})}. Both raw observations and segmentation masks are provided to the VLM in this process; and (3) a dataset is collected using the VLM-generated code to train a lightweight vision model (the \textit{Extractor Module}) as a computationally efficient substitute \edit{(line 5 and line 6 in Algorithm~\ref{alg:itrace}). Step 3 is crucial because VLM-generated code depends on image segmentation, which is prohibitively slow and non-batchable when performed at every RL step, resulting in impractical training times.}


\subsection{Deduce features}

To extract intelligible features from an unseen environment that could be used to train an effective RL policy, we need to first determine what features are useful to model decision-making (line 3 in Algorithm~\ref{alg:perception}). In this step, we provide the VLM with:

\begin{itemize}
    \item \( \mathcal{D} \): Domain knowledge limited to basic information, obtainable via foundation model queries or agentic exploration (Sec.~\ref{sec:future_work}). 
    \item \( \mathcal{I}_0 \): A diverse list of example image observations from the environment.
\end{itemize}

The VLM identifies key semantic features critical for decision-making, represented as two distinct sets:

\begin{itemize}
    \item \( \mathcal{F}_s = \{f_1, f_2, \dots, f_p\} \): Features that can derived from a single image frame.
    \item \( \mathcal{F}_m = \{f_{p+1}, f_{p+2}, \dots, f_n\} \): Features requiring multiple frames (e.g., velocity, acceleration, tendency of change). Note that \( \mathcal{F}_m \) may not be needed in every environment.
\end{itemize}

The process can be formalized in Eq.~\ref{eq:vlm_step_1}:

\par\nobreak{\parskip0pt \noindent
\begin{equation}
    \mathcal{F} = \mathcal{F}_s \cup \mathcal{F}_m, \quad \mathcal{F} = \mathcal{M}(\mathcal{D}, \mathcal{I}_0).
\label{eq:vlm_step_1}
\end{equation}}

\vspace{-0.5em}
For example, in the Skiing environment, as deduced by the VLM, \( \mathcal{F}_s\) includes the positions of the next gate (left pole x, right pole x, and y), the closest tree (x, y) and the skier's x-position,
while \( \mathcal{F}_m\) includes the horizontal speed of the skier.

We instruct the VLM to give a \texttt{"temporal\_context"} boolean field for each feature, enabling it to automatically distinguish between static and temporal features and assign them to \( \mathcal{F}_s\) or \( \mathcal{F}_m\). While \( \mathcal{F}_m\) may be unnecessary in some environments, it is essential in others. Returning to the Skiing example, the skier must actively tilt left and right to maintain speed. Without knowledge of the skier’s motion direction, it is ambiguous which way the skier should tilt. We allow the VLM to determine whether temporal context is required for effective decision-making in a given environment. 

\vspace{-0.5em}
\subsection{Generate feature extraction code}
\vspace{-0.1em}
In this step, we query the VLM to generate feature extraction code, denoted by \( \mathcal{G} \), to retrieve the identified set of features \(\mathcal{F}\) from one or multiple image-segmentation pairs \( (\mathcal{I}, \mathcal{S})\). 
Recognizing that hallucination is a well-documented challenge in generative VLM models \cite{liu2024surveyhallucinationlargevisionlanguage}, we use segmentation to provide accurate spatial information and ground the generated extraction code in visual data. This strategy aligns with common practices in the field \cite{peng2024preference,yang2023finegrained}. 
The visual segmentation \( \mathcal{S} \) is computed using Segment Anything Model (SAM) \cite{Kirillov_2023_Segment_Anything}, denoted by \(Seg\), such that \( \mathcal{S} = \text{Seg}(\mathcal{I}) \). We use the ViT-H SAM model with its default parameters and no modifications to ensure the generalizability of our approach.

\textbf{Example Segmentations Labeling}: 
While segmentation improves spatial grounding, its interpretation by the VLM remains prone to hallucination.
To ensure the generated code generalizes across diverse environments, the VLM must be exposed to a representative set of example images.
However, providing multiple segmentation JSONs at once can cause the VLM to conflate objects across images, consistent with known failures of VLMs to maintain object identity across contexts~\cite{li2025mihbench}.
To address this, we first prompt the VLM to label a small set of segmentations associated with example frames, processing one image at a time. An example of this labeling process is shown in Fig.~\ref{fig:sub2:vlmlabel}, where all segment labels are generated by the VLM based on raw segmentation masks.

Following this, depending on whether the feature set requires temporal context, the VLM is instructed to take one or multiple image-segmentation pairs \( (\mathcal{I}, \mathcal{S}) \). The labeled segmentations serve as reference examples, helping the VLM infer the most likely attribute values of relevant objects and formulate logic to filter out noisy segments.
The code generation process is described in line 4-9 in Algorithm~\ref{alg:perception}), where the VLM first generates code to extract values from a single image–segmentation pair \( (\mathcal{I}, \mathcal{S}) \), and then generates code \( \mathcal{G}_m \), extending \( \mathcal{G}_s \) to handle sequences of image-segmentation pair \(\{(I_{t-i}, S_{t-i}), \dots, (I_t, S_t)\}\).

\begin{algorithm}[tb]
\caption{iTRACE Overall Framework}
\label{alg:itrace}
\begin{algorithmic}[1]
\STATE {\bfseries Input:} Environment $\mathcal{E}$, VLM $\mathcal{M}$, RL algorithm (PPO), horizon $H$
\STATE // Call perception stack
\STATE $F \leftarrow$ \textsc{DeduceFeatures}($\mathcal{M}, \mathcal{E}$) 
\STATE $G \leftarrow$ \textsc{GenerateExtractionCode}($\mathcal{M}, F$) 
\STATE $\mathcal{T} \leftarrow$ \textsc{CollectDataset}($\mathcal{E}, G$) 
\STATE $\phi \leftarrow$ \textsc{TrainExtractorModule}($\mathcal{D}$) 
\STATE // Call RL stack
\STATE $\pi_\theta \leftarrow$ \textsc{TrainICT}($\mathcal{M}, \mathcal{E}$) 
\STATE {\bfseries Return:} trained ICT policy $\pi_\theta$
\end{algorithmic}
\end{algorithm}

\begin{algorithm}[tb]
\caption{Perception Stack: Automated Feature Extraction}
\label{alg:perception}
\begin{algorithmic}[1]
\STATE {\bfseries Input:} Environment $\mathcal{E}$, VLM $\mathcal{M}$, Domain $\mathcal{D}$
\STATE Sample example images $\{I_i\}_{i=1}^n$ from $\mathcal{E}$

\CommentColor  {Step 1: Deduce features}
\STATE $F \leftarrow \mathcal{M}_{\text{Extract}}(D, \{I_i\})$

// $\mathcal{M}$ parses $F$ into static $F_s$ and temporal $F_m$.

\CommentColor  {Step 2: Generate extraction code}
\STATE $G_s \leftarrow \mathcal{M}_{\text{Generate}}(F_s, \{(I_i, \text{Seg}(I_i))\})$
\IF{$F_m \neq \emptyset$}
  \STATE $G_m \leftarrow \mathcal{M}_{\text{Generate}}(F_m, G_s)$
  \STATE $G \leftarrow (G_s, G_m)$
\ELSE
  \STATE $G \leftarrow G_s$
\ENDIF

\CommentColor  {Step 3: Train extractor module}
\STATE Collect dataset $\mathcal{T} = \{(I_t, G(I_t))\}$ via exploration
\STATE Train lightweight extractor $\phi$ on $\mathcal{T}$
\STATE {\bfseries Return:} Extractor module $\phi$
\end{algorithmic}
\end{algorithm}

\begin{algorithm}[tb]
\caption{RL Stack: Training Interpretable Control Tree}
\label{alg:rl}
\begin{algorithmic}[1]
\STATE {\bfseries Input:} Extractor module $\phi$, environment $\mathcal{E}$, horizon $H$, PPO parameters
\STATE Initialize ICT policy $\pi_\theta$
\FOR{episode $=1, \dots, N$}
  \STATE Reset $\mathcal{E}$, observe raw frame $I_0$
  \FOR{t = 1 to H}
    \STATE $f_t \leftarrow \phi(I_t)$ \hfill // semantic features
    \STATE $a_t \sim \pi_\theta(f_t)$
    \STATE Apply $a_t$ in $\mathcal{E}$, receive $I_{t+1}, r_t$
  \ENDFOR
  \STATE Update $\pi_\theta$ via PPO on collected $(f_t, a_t, r_t)$
\ENDFOR
\STATE {\bfseries Return:} trained ICT policy $\pi_\theta$
\end{algorithmic}
\end{algorithm}

\begin{figure*}[htbp]
  \centering
  \begin{subfigure}[b]{0.25\textwidth}
    \includegraphics[width=\textwidth]{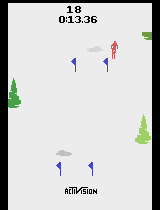}
    \caption{Original image observation.}
    \label{fig:sub1:original}
  \end{subfigure}
  \hfill
  \begin{subfigure}[b]{0.25\textwidth}
    \includegraphics[width=\textwidth]{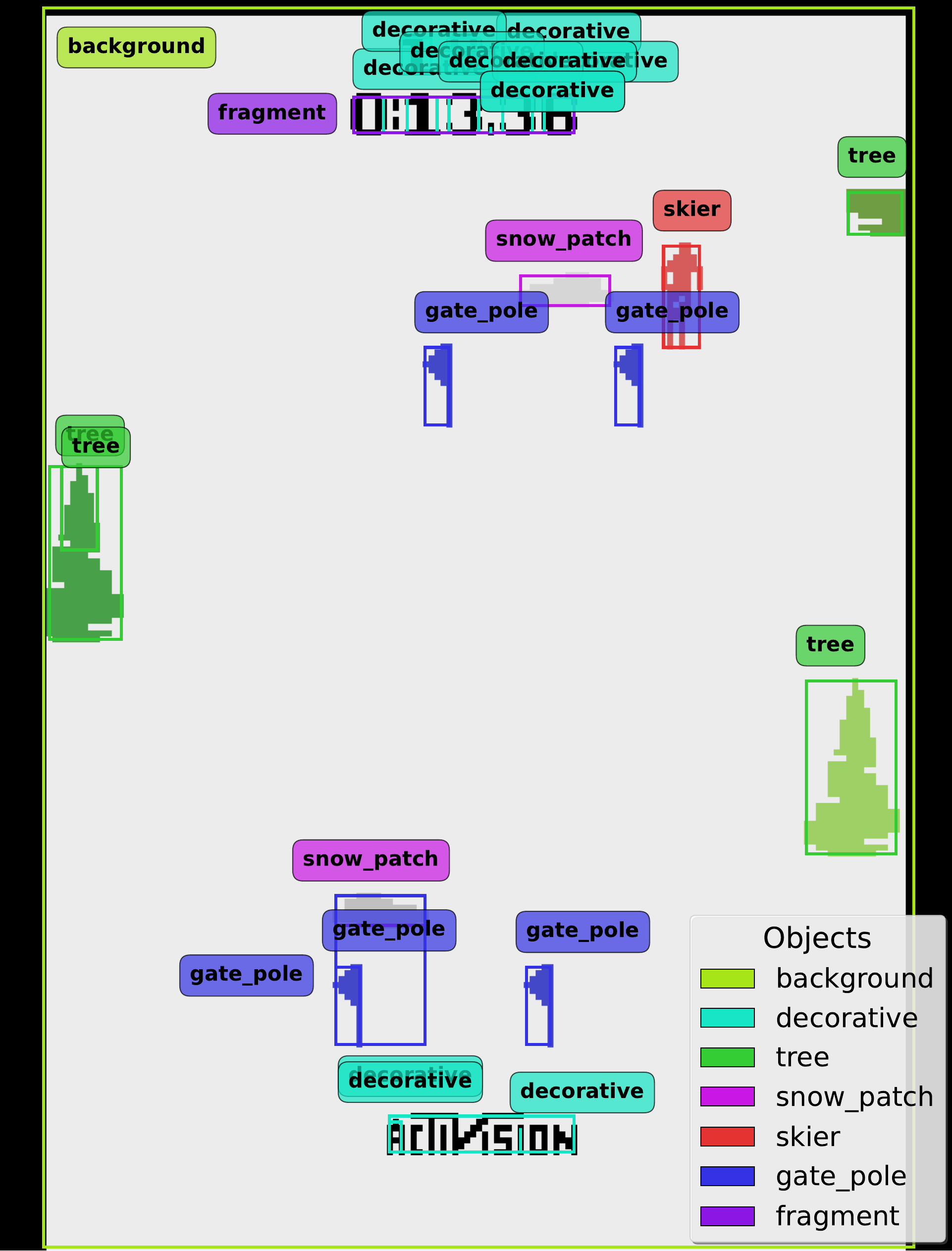}
    \caption{VLM-labeled segmentation.}
    \label{fig:sub2:vlmlabel}
  \end{subfigure}
  \hfill
  \begin{subfigure}[b]{0.25\textwidth}
    \includegraphics[width=\textwidth]{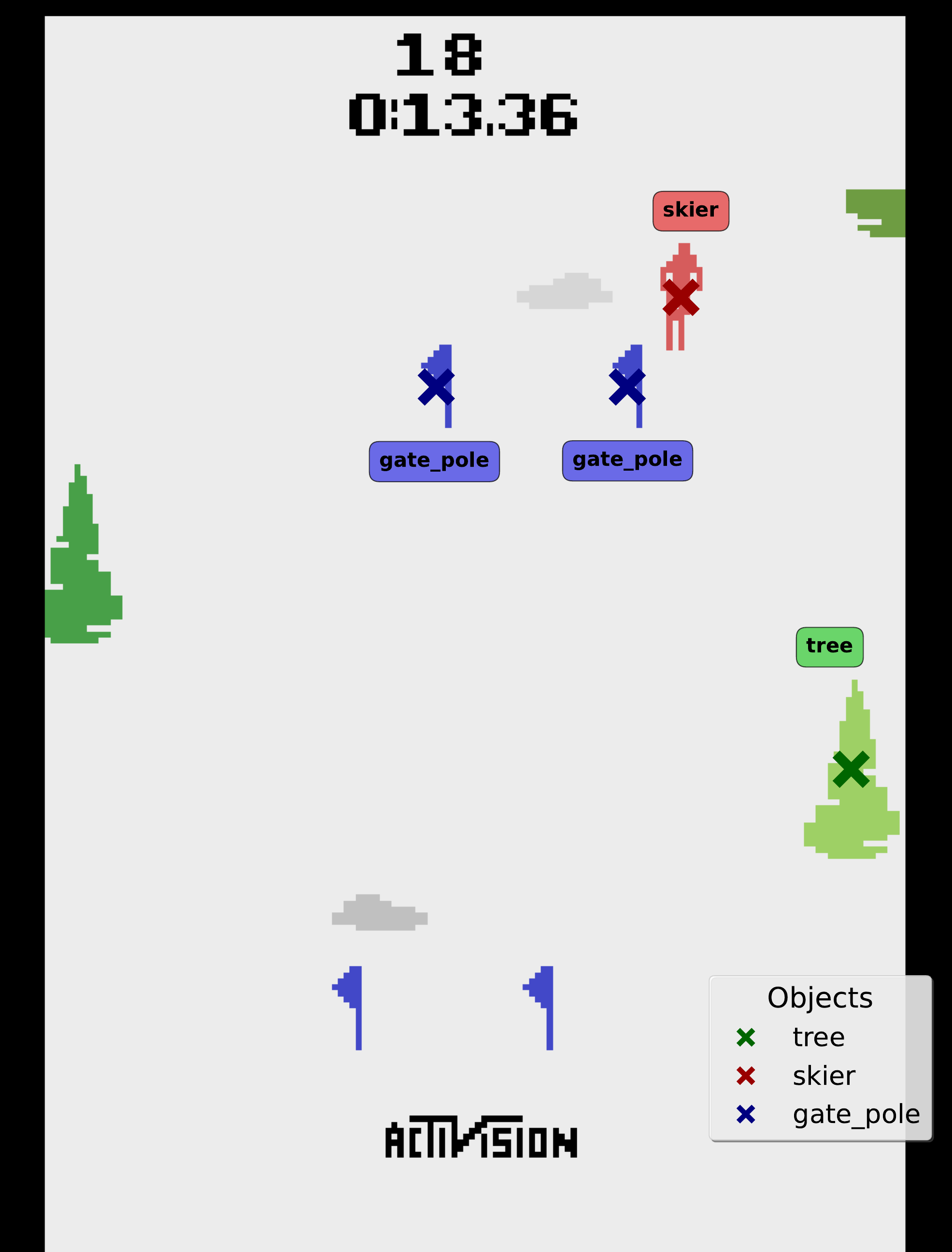}
    \caption{VLM-extracted features.}
    \label{fig:sub3:vlmextract}
  \end{subfigure}
  \caption{An example of how raw image observations are processed in our pipeline. 
  (a) Raw image observations are segmented using SAM. 
  (b) A VLM assigns semantic labels to segments in a small set of example images. 
  (c) A VLM then generates code that extracts target features from segmentation masks, applicable to all images. 
  Step (b) is only performed on the example set to guide code generation.}
  \label{fig:image_segmentation_example}
\end{figure*}

\vspace{-0.5em}
\subsection{Dataset generation and training the extractor module.}

In this step, we address the inefficiencies of the VLM-generated feature extraction code for single-image frames, $\mathcal{G}_s$, by training an lightweight Extractor Module as its replacement. While $\mathcal{G}_s$ can be executed directly on the dataset, the processing pipeline is inherently non-batchable, as it handles image segmentations independently. This non-batchability severely impacts downstream RL training. 

We execute \( \mathcal{G} \) on trajectories collected from the environment to generate the feature dataset. For each trajectory, a series of images, \( \mathcal{T} = \{I_1, I_2, \dots, I_k\} \) is generated, with corresponding visual segmentations \( \mathcal{S}_t = \text{Seg}(I_t) \) for each \( t \). The extractor code \( \mathcal{G} \) processes each image-segmentation pair \( (I_t, S_t) \) independently to extract features \( \mathbf{f}_t = \{f_1, f_2, \dots, f_n\} \) (line 11 in Algorithm~\ref{alg:perception}). 

The dataset, consisting of image-feature pairs \( \{(\mathcal{I}_t, \mathbf{f}_t)\} \), is then used to train an Extractor Module as a batch-operable and efficient replacement to \( \mathcal{G}_s \) (line 12 in Algorithm~\ref{alg:perception}). The trained model predicts semantic features directly from raw images, enabling significantly faster feature extraction during RL training. In our experiments, we use a CNN-based Extractor Module due to its low computational overhead in simulated environments. 
\polina{While the framework is compatible with any lightweight vision model, including fine-tuned ViTs or task-specific architectures, CNNs provide an effective and widely validated baseline that balances accuracy, efficiency, and robustness. The execution time comparison between VLM-generated feature extraction code and our CNN-based Extractor Module is demonstrated in Table \ref{tab:cnn_performance_time}.}

\begin{table}[h]
\centering
\caption{Runtime speedup on Skiing. Similar results achieved on other domains.}
\label{tab:cnn_performance_time}
\begin{tabular}{cccc}
\toprule
\textbf{Batch} & \textbf{VLM Code} & \textbf{CNN} & \textbf{Speedup} \\
\midrule
8   & 19.056  & 0.001 & 13831.3$\times$ \\
16  & 38.288  & 0.002 & 17325.6$\times$ \\
32  & 76.212  & 0.004 & 19935.0$\times$ \\
64  & 151.836 & 0.007 & 21692.9$\times$ \\
128 & 308.468 & 0.014 & 22772.4$\times$ \\
\bottomrule 
\end{tabular}
\end{table}

If some features require temporal context (i.e., \( \mathcal{F}_m \neq \emptyset \)), \( \mathcal{G}_m \), consisting of simple mathematical operations on the outputs of \( \mathcal{G}_s \), is adapted to utilize the Extractor Module in place of \( \mathcal{G}_s \). 
For example, in the Skiing environment, \( \mathcal{G}_m \) calculates skier horizontal speed by subtracting the positions returned by \( \mathcal{G}_s \)

\vspace{-0.5em}
\section{Tree-based RL training} 

\edit{Leveraging the semantic features extracted from the raw visual inputs courtesy of the perception stack, we now focus on training an interpretable RL model that maps semantic features to action outputs}, as shown in Fig.~\ref{fig:framework_overview}. The policy network takes these semantic features as input and predicts a distribution over actions, from which an action is sampled and applied to the controlled agent at each time step. 

Unlike traditional RL algorithms that use black-box models as the policy network, we use an Interpretable Control Tree (ICT) \edit{adapted from Interpretable Continuous Control Tree (ICCT) proposed by Paleja et al. ~\cite{paleja2023interpretable}} as our policy network to achieve transparent decision-making and human-aligned reasoning while remaining differentiable for end-to-end optimization. 
Unlike ICCT, ICT supports both discrete \emph{and} continuous actions without re-architecture, which helps generalizability. In the discrete case, leaf nodes in ICT output the possibility of taking each action as a Probability Mass Function (PMF). If there are six possible actions in the discrete action space, then there will be six sparse linear functions of the input features followed by a softmax function at each leaf node. And for the continuous action space, leaf nodes output the Probability Density Function (PDF) over possible actions. The detailed RL training is listed in Algorithm~\ref{alg:rl}

\subsection{Interpretable control tree}

Each decision node in ICT determines the routing of input features, $\mathbf{x}$, to sparse linear controllers at the leaf nodes. 

Each decision node, \(i\), is parameterized by weights \(\mathbf{w}_i\), a bias \(b_i\), and a steepness parameter \(\alpha\) to output a fuzzy Boolean in $[0,1]$, as defined in   Eq.~\ref{eq:icct_simplified_decision_function}.
\par\nobreak{\parskip5pt \noindent
\begin{equation}
    y_i = \sigma\left(\alpha \left(w_i^k x^k - b_i \right)\right)
\label{eq:icct_simplified_decision_function}
\end{equation}}The most important feature, $k$, of node, $i$, is selected in Eq.~\ref{eq:icct_crispification} by identifying the weight with the largest magnitude 
\par\nobreak{\parskip5pt \noindent
\begin{equation}
    k = \arg\max_j \left| w_i^j \right|
\label{eq:icct_crispification}
\end{equation}}Since the \(\arg\max\) operator is non-differentiable, we approximate this function with the Gumbel-Softmax ~\cite{jang2016categorical}, as per Eq.~\ref{eq:gumbel_softmax}, where \( g_j \sim \text{Gumbel}(0,1) \) and \( \tau \) is the temperature.
\par\nobreak{\parskip5pt \noindent
\begin{equation}
    f(w_i)^k = \frac{\exp \left(\frac{w_i^k + g_k}{\tau} \right)}
    {\sum_j \exp \left(\frac{w_i^j + g_j}{\tau} \right)}
\label{eq:gumbel_softmax}
\end{equation}}During inference, a hard threshold is applied to enforce discrete decision-making using only one feature per node, as per Eq.~\ref{eq:hard}, where \(\ind\) is the indicator function.
\par\nobreak{\parskip5pt \noindent
\begin{equation}
    y_i = \ind(\alpha(w_i^k x^k - b_i) > 0)
    \label{eq:hard}
\end{equation}}During training, gradients flow through discrete decisions using the straight-through trick~\cite{bengio2013estimating}. This approach ensures each decision node remains interpretable while allowing end-to-end gradient-based learning.

Each leaf node, $\ell_d$, in the ICT is parameterized by a set of sparse linear controllers, as formulated in Eq.~\ref{eq:icct_sparse_leaf}, where $\mathbf{\beta}_d$ is a sparsely activated weight vector, $\phi_d$ is a bias term, and $\mathbf{u}_d$ is a $k$-hot encoding vector that enforces sparsity, computed in Eq.~\ref{eq:icct_leaf_feature_selection}. The per-leaf selector weight, $\theta_d$, learns the relative importance of different input features. The function $h(\cdot)$ provides a differentiable approximation of the $k$-hot function by assigning a value of 1 to the top $k$ elements in $|\mathbf{\theta}_d|$ (based on magnitude) and 0 to the rest. Differentiability is preserved using the straight-through trick~\cite{bengio2013estimating}.
\par\nobreak{\parskip5pt \noindent
\begin{equation}
    PMF / PDF = (\mathbf{u}_d \circ \mathbf{\beta}_d)^\top (\mathbf{u}_d \circ \mathbf{x}) + \mathbf{u}_d^\top \phi_d
\label{eq:icct_sparse_leaf}
\end{equation}}\par\nobreak{\parskip0pt \noindent
\begin{equation}
    \mathbf{u}_d = h\left(\mathrm{softmax}\left(|\mathbf{\theta}_d|\right)\right),
\label{eq:icct_leaf_feature_selection}
\end{equation}}

\section{Experiments}

\subsection{Environment}

We evaluate iTRACE in three diverse environments chosen to ensure the VLM performs genuine reasoning rather than relying on memorized knowledge~\cite{carlini2023quantifyingmemorizationneurallanguage}: Atari Skiing, a pixel-art game; BabyAI-GoToRedBall, a symbolic grid world for navigation; and OpenAI Gym Highway, a 2D driving environment. While BabyAI and Highway provide default feature vectors, these describe the full environment state rather than the task-relevant semantic features required for decision making. The Atari Skiing environment provides raw image state spaces without any pre-defined feature vectors or feature sets. 
BabyAI and Highway offer default feature vectors, but these vectors describe the full state of the environment rather than focusing on the task-relevant semantic features needed for decision making.

\vspace{-0.3em}
\subsection{Feature extraction}
\subsubsection{VLM deducing key semantic features.}
For this step, we utilize \texttt{ChatGPT-4o-latest}, one of the current leading models
in language reasoning.
Across all three environments, the VLM consistently produces a deterministic set of semantic features for decision-making, achieving performance comparable to human reasoning. For all three environments, we query the VLM 50 times and use majority vote to get the most prominent answer from the VLM. A complete set of those features are included in Table~\ref{tab:feature_schemas}.

\begin{table}[h]
\setlength{\tabcolsep}{0.5pt}
\centering
\caption{Feature schemas for BabyAI RedBall, Highway, and Skiing environments. 
Type: C = Continuous, B = Binary. Temporal: T = True, F = False.}
\label{tab:feature_schemas}
\begin{tabularx}{\columnwidth}{c@{\hspace{6pt}} >{\scriptsize\ttfamily\arraybackslash}X c c}
\toprule
& \textbf{Feature} & \textbf{Type}\hspace{0.2em} & \textbf{Temp.} \\
\midrule
\multirow{3}{*}{\rotatebox{90}{\parbox{3.4em}{\scriptsize \centering BabyAI}}}
  & \texttt{red\_ball.\{x,y\}\_position} & C & F \\
  & \texttt{red\_ball.is\_visible} & B & F \\
  & \texttt{obstacle\_\{forward,left,right\}.is\_blocked} & B & F \\
\midrule
\multirow{6}{*}{\rotatebox{90}{\parbox{3.4em}{\scriptsize \centering Highway}}}
  & \texttt{agent.\{x,y\}\_position} & C & F \\
  & \texttt{agent.x\_velocity} & C & T \\
  & \texttt{lane\_existence.\{upper,lower\}} & B & F \\
  & \texttt{vehicle\_ahead.\{x\_position,x\_velocity\}} & C & F/T \\
  & \texttt{vehicle\_upper\_lane.\{x\_position,x\_velocity\}} & C & F/T \\
  & \texttt{vehicle\_lower\_lane.\{x\_position,x\_velocity\}} & C & F/T \\
\midrule
\multirow{5}{*}{\rotatebox{90}{\parbox{3.4em}{\scriptsize \centering Skiing}}}
  & \texttt{next\_\{left,right\}\_pole.x\_position} & C & F \\
  & \texttt{next\_pole.y\_position} & C & F \\
  & \texttt{next\_tree.\{x,y\}\_position} & C & F \\
  & \texttt{skier.x\_position} & C & F \\
  & \texttt{skier.x\_velocity} & C & T \\
\bottomrule
\end{tabularx}
\normalsize
\end{table}

\subsubsection{VLM generating Feature Extraction code.}

For this step, we utilize \texttt{claude-3.5-sonnet-20241022}, one of the leading models in code generation. To address common challenges with VLM-generated code, such as non-functionality or runtime errors, we query the VLM 50 times using the same prompt (temperature = 0 for reproducibility) and select the most reliable feature extraction code through majority voting.

\begin{figure}[htbp]
  \centering
  \begin{subfigure}[b]{0.15\textwidth}
    \includegraphics[width=\textwidth]{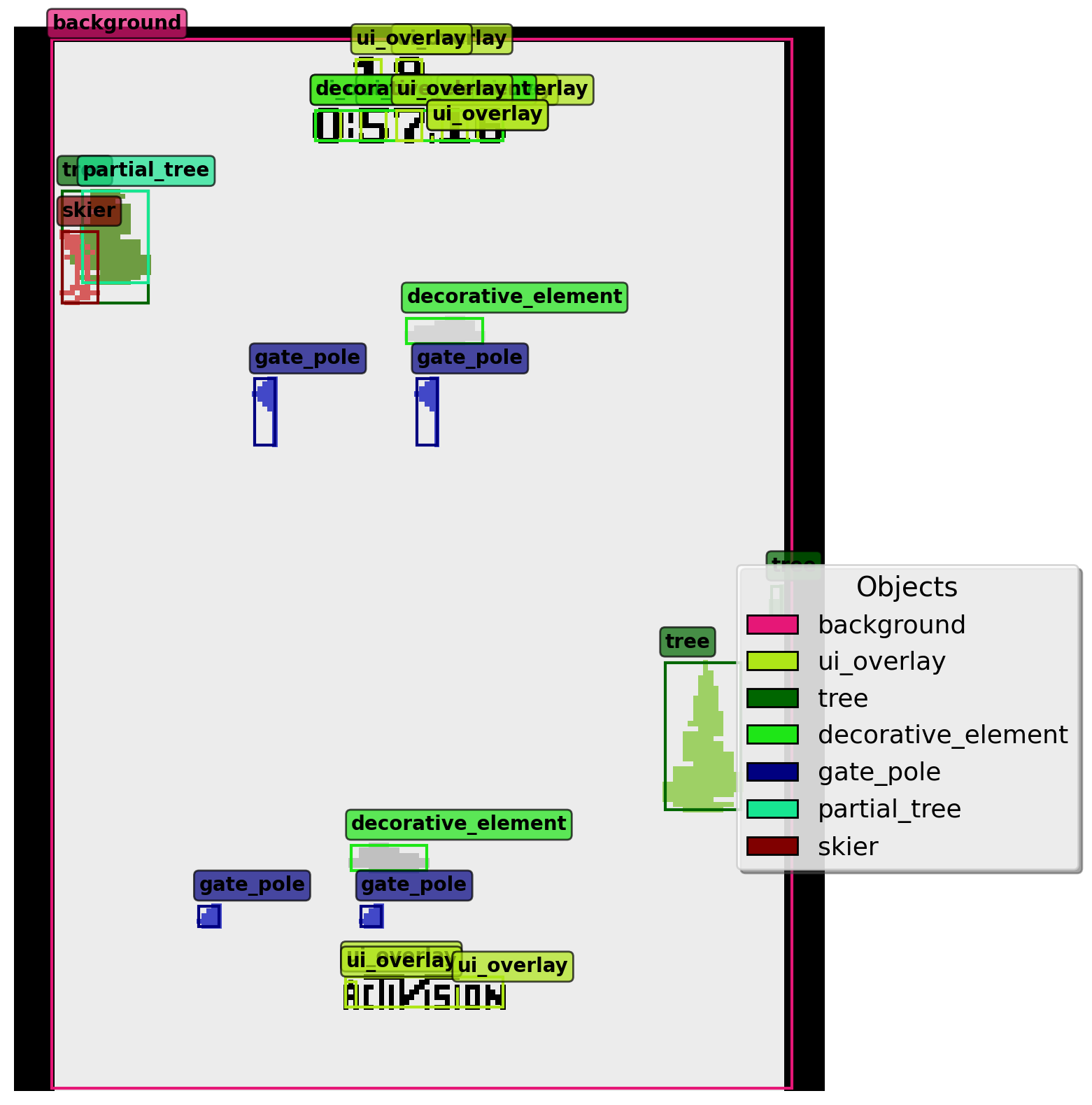}
    \caption{Skiing Exp. 1.}
    \label{fig:sub1:original}
  \end{subfigure}
  \hfill
  \begin{subfigure}[b]{0.15\textwidth}
    \includegraphics[width=\textwidth]{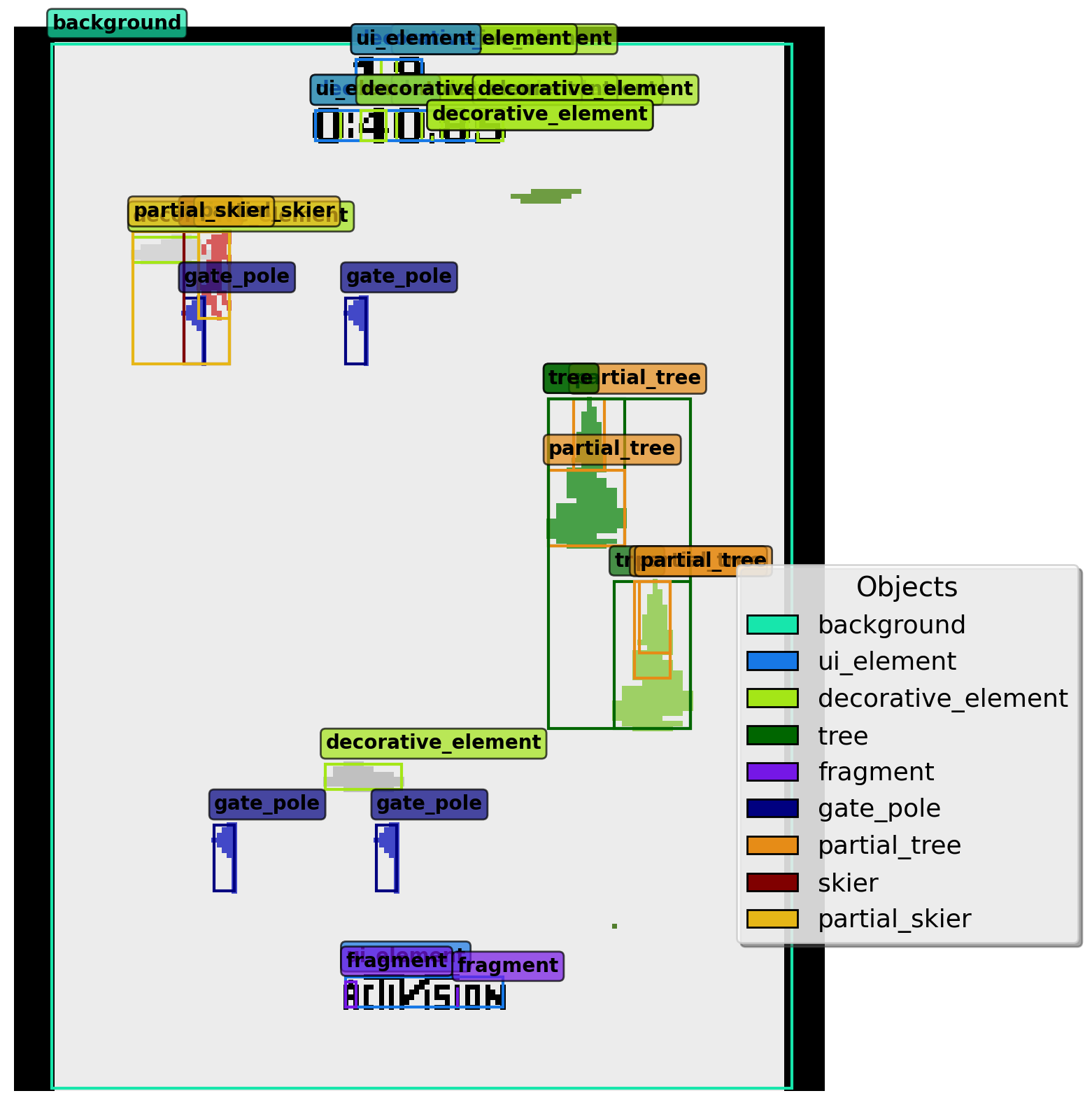}
    \caption{Skiing Exp. 2.}
    \label{fig:sub2:vlmlabel}
  \end{subfigure}
  \hfill
  \begin{subfigure}[b]{0.15\textwidth}
    \includegraphics[width=\textwidth]{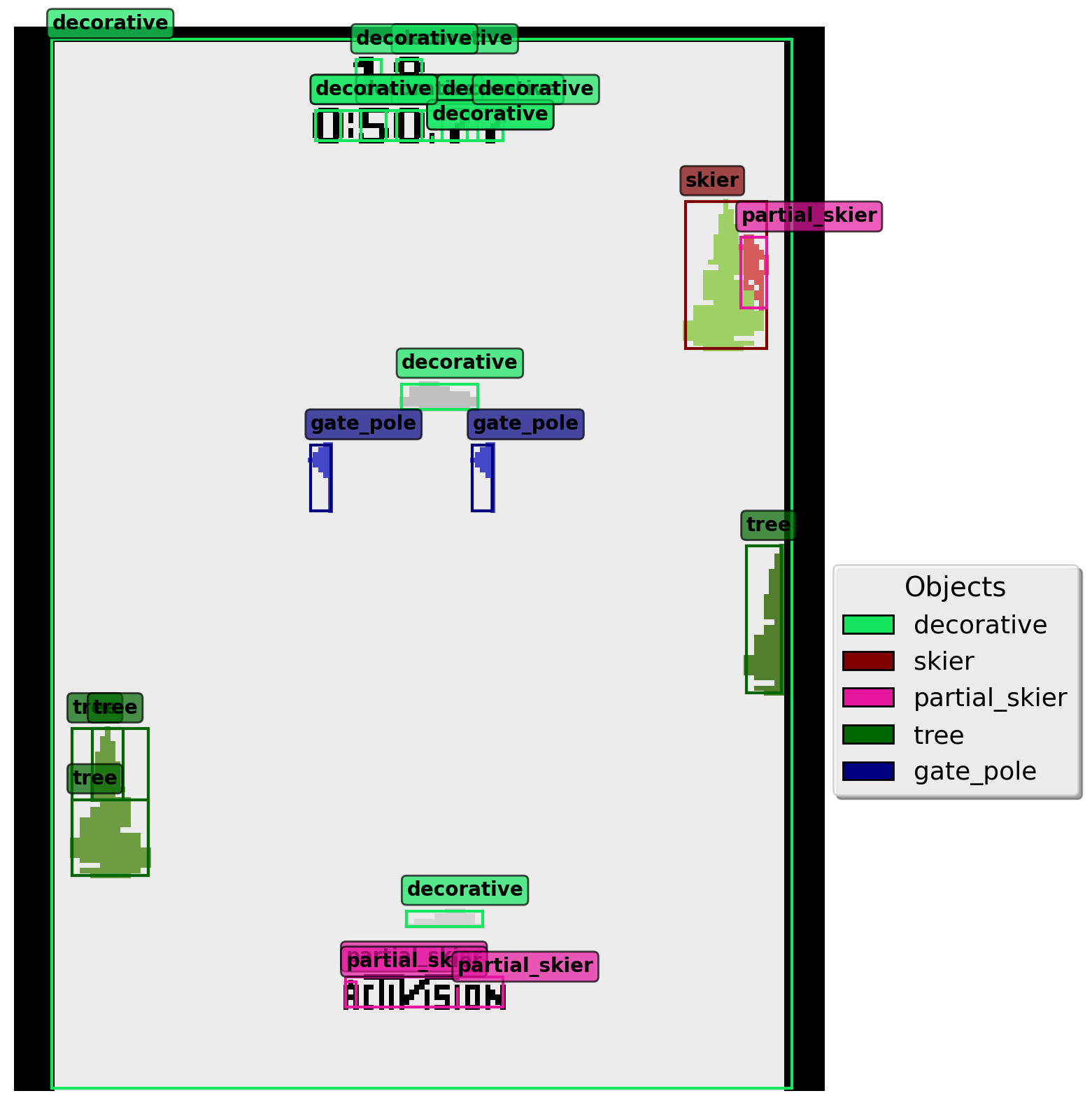}
    \caption{Skiing Exp. 3.}
    \label{fig:sub3:vlmextract}
  \end{subfigure}
  \caption{Randomly selected segmentation examples we provide to the VLM for domain Skiing. Example 1 is good while 2 and 3 are imperfect. Example 2 identifies the skier as a partial skier and a noisy segment as skier. Example 3 identifies both the actual skier on the right and decoration elements at the bottom as partial skier.}
  \label{fig:image_segmentation_example_skiing}
\end{figure}

Since code generation depends on segmentation quality $\mathcal{S}$, we further evaluate segmentation performance.
Among the evaluated environments, Skiing presented the greatest difficulty for VLM-based segmentation labeling, and we therefore selected it for detailed analysis. On 100 sampled images, segmentation achieves high overall accuracy (Table~\ref{tab:confusion_ratio}) but includes imperfect cases where the skier is partially or entirely missed (Fig.~\ref{fig:image_segmentation_example_skiing}). Despite such errors present in our in-context-learning examples (Fig.~\ref{fig:image_segmentation_example_skiing}), the generated code remains effective, demonstrating the robustness of our framework as long as a subset of segments is correctly labeled.





The generated codes are evaluated on a randomly collected dataset of 100 images and filtered based on the following criteria: (1) the code must compile and execute successfully and (2) it must produce diverse feature outputs rather than constant or repetitive values across the dataset. Candidates producing identical outputs are grouped into clusters, and the largest cluster is chosen as the final feature extraction code.

\begin{table}[h]
\centering
\caption{Confusion ratios per feature concept}
\label{tab:confusion_ratio}
\begin{tabular}{c|c|c|c}\toprule
& Skier & Pole & Tree \\
\midrule
\begin{tabular}{@{}cc@{}} \cellcolor{black!24}{TP} & FN \\FP & \cellcolor{black!24}{TN} \end{tabular} & 




\begin{tabular}{@{}cc@{}}\cellcolor{black!24}9.21 & \cellcolor{black!0}1.82 \\\cellcolor{black!0}00.21 & \cellcolor{black!24}{88.76}\end{tabular} & 

\begin{tabular}{@{}cc@{}}\cellcolor{black!24}27.73 & \cellcolor{black!0}00.11 \\\cellcolor{black!0}00.43 & \cellcolor{black!24}{71.73}\end{tabular} & 

\begin{tabular}{@{}cc@{}}\cellcolor{black!24}58.03 & \cellcolor{black!0}3.11 \\\cellcolor{black!0}00.1071 & \cellcolor{black!24}38.76\end{tabular} 

\\
\bottomrule\end{tabular}
\end{table}

\subsubsection{Extractor module performance analysis}

We train an CNN based Extractor Module to resolve the slowdown caused by segmentation-based analysis.
For each environment, we train an CNN on 7,000 images collected from diverse trajectories, validate it on 2,000 images, and test it on 1,000 images. 
As shown in Table \ref{tab:cnn_performance_error}, compared to the VLM-generated ground truth, our models achieve high accuracy (typically >0.95) on most binary features, while MSE remains low (<0.005) for the majority of continuous features, with one exception where error is notably higher.
These results confirm that a lightweight vision model can serve as a reliable and computationally efficient replication for VLM generated extraction code $\mathcal{G}_s$.

\begin{table}[t]
\centering
\caption{Performance of the Extractor Module across environments, compared with ground truth from VLM-generated code. Continuous components are evaluated with MSE; boolean components are evaluated with accuracy. Dashes indicate vacant entries due to varying feature lengths and types across environments. Full feature descriptions are provided in Table~\ref{tab:feature_schemas}.}
\label{tab:cnn_performance_error}
\resizebox{\columnwidth}{!}{%
\begin{tabular}{@{}lccccccc@{}}
\toprule
\textbf{Environment} & \multicolumn{7}{c}{\textbf{Feature Vector Indices (1--7)}} \\
\cmidrule(lr){2-8}
 & \textbf{1} & \textbf{2} & \textbf{3} & \textbf{4} & \textbf{5} & \textbf{6} & \textbf{7} \\
\midrule
\textbf{Skiing (MSE)}       & 0.00023 & 0.00025 & 0.0020 & 0.0039 & 0.0024 & 0.00034 & -- \\
\midrule
\textbf{BabyAI (MSE)}       & 0.0001 & 0.0001 & -- & -- & -- & -- & -- \\
\textbf{BabyAI (Acc)}       & -- & -- & 1.000 & 0.979 & 0.975 & 0.972 & -- \\
\midrule
\textbf{Highway (MSE)}      & 0.0000 & 0.0000 & -- & -- & 1.153 & 0.0268 & 0.0209 \\
\textbf{Highway (Acc)}      & -- & -- & 0.988 & 0.948 & -- & -- & -- \\
\bottomrule
\end{tabular}%
}
\end{table}

\vspace{-0.5em}
\subsection{\edit{Baselines}}

We train ICT, our proposed interpretable and verifiable method that learns a tree structure with RL, alongside several baseline models across all environments. Below is an explanation of each baseline model used in the evaluation:

\begin{itemize}
    \item \textbf{CNN}: An uninterpretable black-box model that maps raw visual inputs to the action space, trained with RL.
    \item \textbf{VLM-ICT}: Our proposed interpretable and verifiable method, using an ICT to connect the semantic feature space with the action space, trained with RL.
    \item \textbf{VLM-ICT (Rand Init)}: Our proposed method, but initialized randomly and not trained with RL.
    \item \textbf{VLM-EQL}: An interpretable neural network that employs simple mathematical functions (e.g., square, cube, sine) to map semantic features to the action space~\cite{sahoo2018learning}.
    \item \textbf{VLM-MLP}: An uninterpretable benchmark that connects a multi-layer perceptron between the semantic feature space and the action space.
    \item \textbf{Random Policy}: A policy that selects actions by randomly sampling from the action space.
\end{itemize}

\begin{figure}[t]
\centering
\begin{subfigure}[t]{\columnwidth}
  \centering
  \includegraphics[width=0.89\columnwidth]{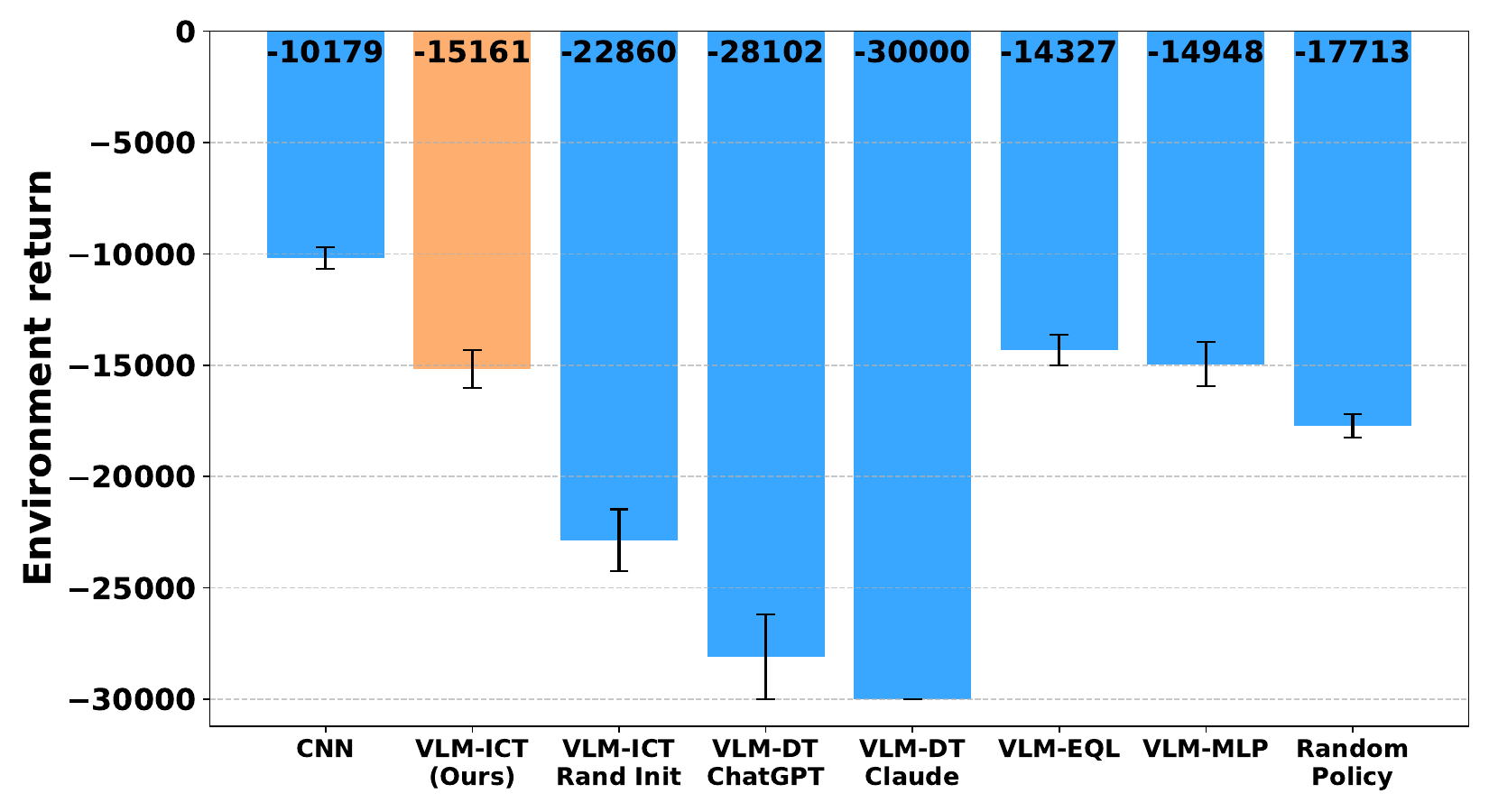}
  \caption{Atari Skiing}
  \label{fig:skiing}
\end{subfigure}

\begin{subfigure}[t]{\columnwidth}
  \centering
  \includegraphics[width=0.89\columnwidth]{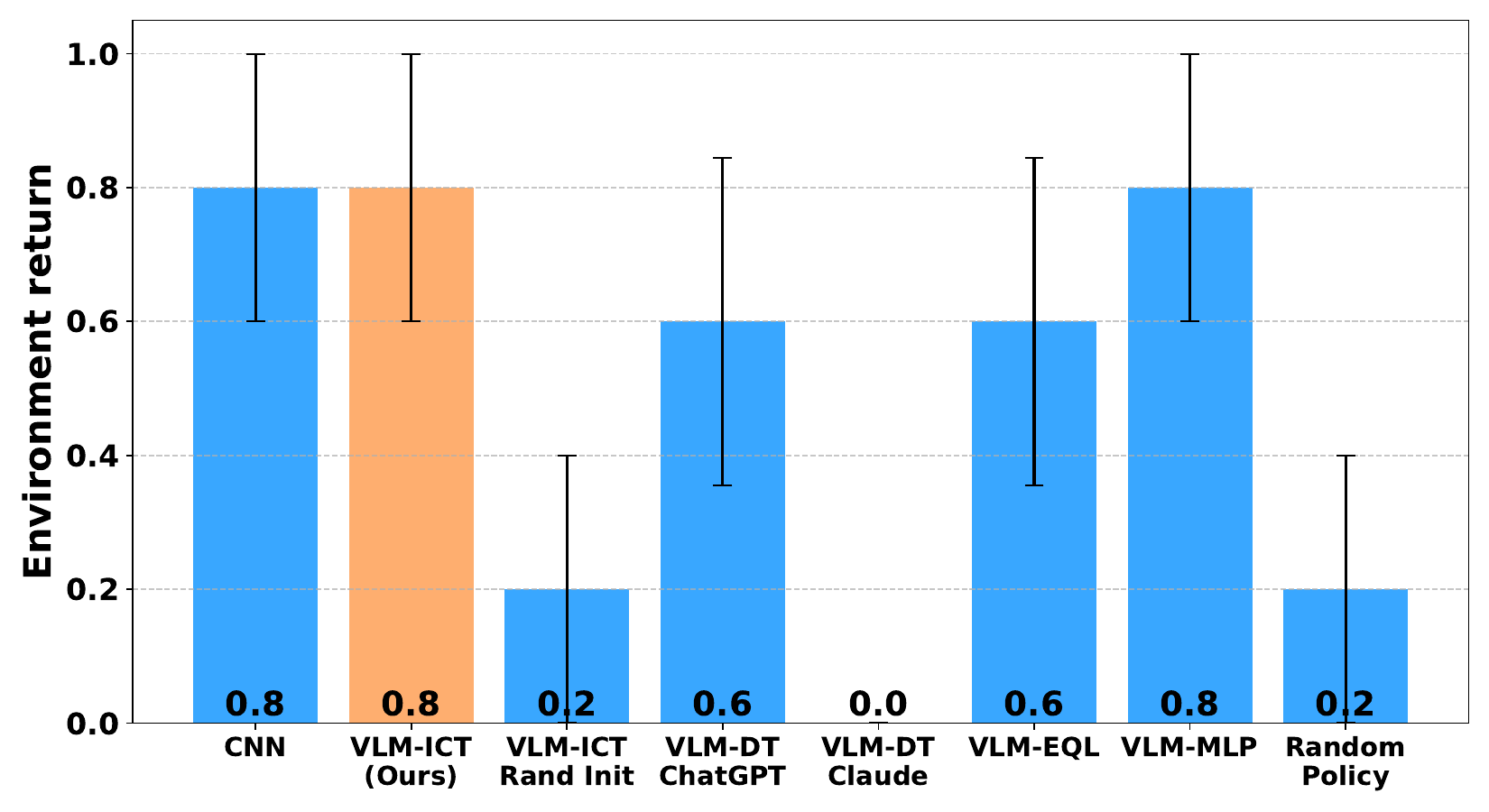}
  \caption{BabyAI: GoToRedBall}
  \label{fig:gotoredball}
\end{subfigure}

\begin{subfigure}[t]{\columnwidth}
  \centering
  \includegraphics[width=0.89\columnwidth]{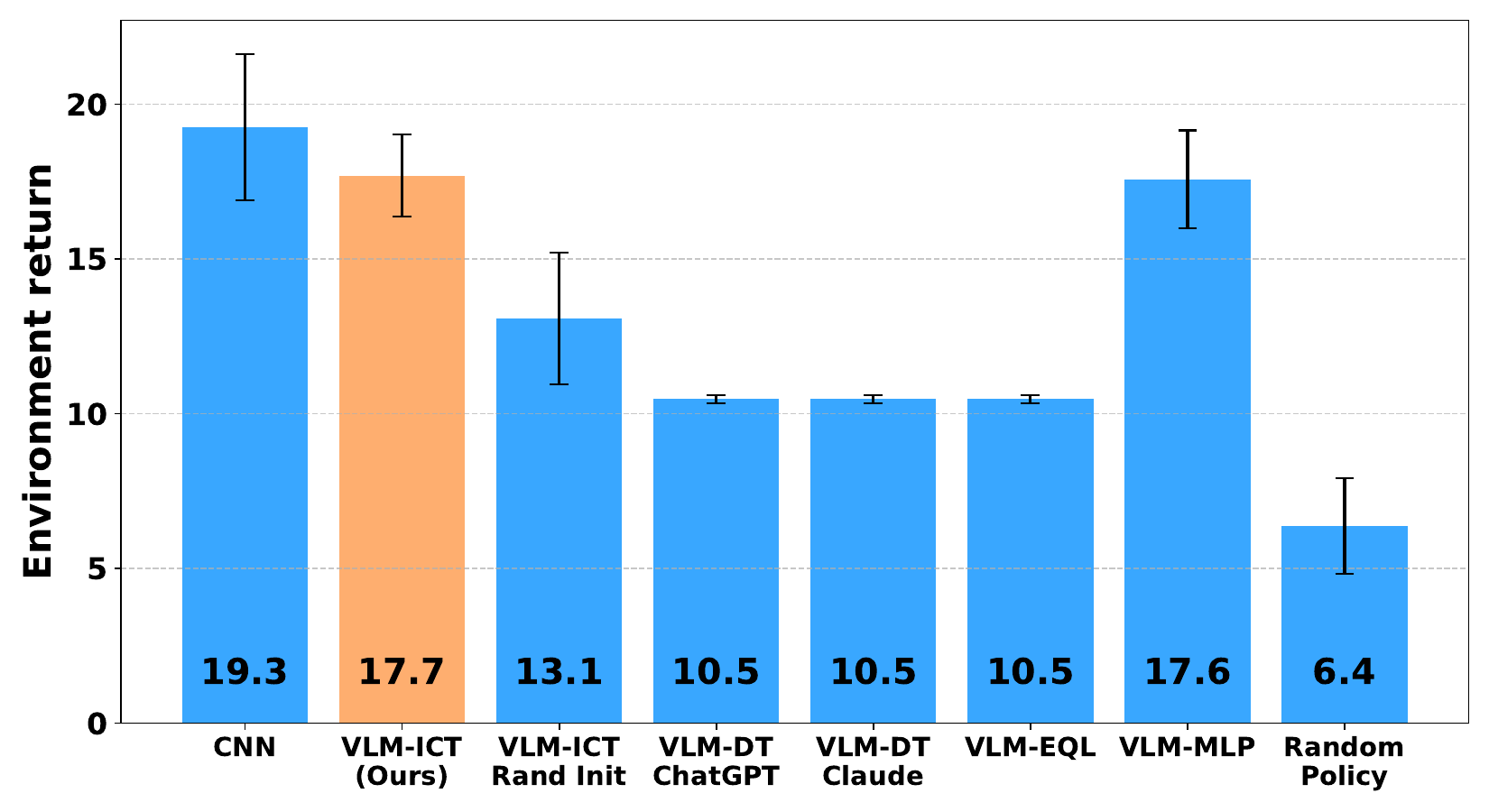}
  \caption{OpenAI Gym Highway}
  \label{fig:highway}
\end{subfigure}

\caption{Results across three domains. Higher scores are better.}
\label{fig:rl_results}
\vspace{-2em}
\end{figure}

In addition to standard learning-based baselines, we also explore whether a Vision-Language Model (VLM) can directly perform action selection via prompting, given access to a semantic feature space it previously constructed. 
We prompt the VLM to synthesize an if-else style decision tree and obtain two variants of this baseline:

\begin{itemize}
\item \textbf{VLM-DT} (\texttt{ChatGPT-4o-latest}): A baseline where ChatGPT-4o generates a heuristic decision tree for action selection.
\item \textbf{VLM-DT} (\texttt{claude-3.5-sonnet}): An equivalent baseline querying Claude 3.5 Sonnet.
\end{itemize}



\subsection{Train results}

We use Proximal Policy Optimization (PPO) as the reinforcement learning algorithm to train our policy~\cite{schulman2017proximal}, with a learning rate of 0.0005 and a discount factor of 0.99. 
During testing, the policy is tested five times using different random seeds, each distinct from those used during training and evaluation. 

Results are shown in Fig.~\ref{fig:rl_results}. \edit{Our results demonstrate the remarkable effectiveness of the proposed framework. As expected, the CNN policy achieves the highest performance across all three environments, benefiting from its status as a black-box model with orders of magnitude more parameters than the interpretable baselines. Notably, our proposed VLM-ICT method consistently reduces this gap. In Atari Skiing, VLM-ICT matches both the interpretable baseline (VLM-EQL) and the black-box policy (VLM-MLP) trained on the same features. In BabyAI and Highway, VLM-ICT outperforms the interpretable baseline (VLM-EQL) and reaches performance comparable to the black-box model (CNN, VLM-MLP). Furthermore, across all domains, our interpretable framework steadily outperforms decision trees generated directly by VLMs (VLM-DT ChatGPT and VLM-DT Claude), demonstrating that these are non-trivial control problems even when the semantic features provided are precisely task-relevant, with no redundant or irrelevant information.}

Building on a semantically meaningful feature set, our ICTs contains nodes directly corresponding to human-understandable concepts. While we do not evaluate interpretability directly, this approach is consistent with prior work where performance serves as a proxy for meaningful representations~\cite{pang-wei2020conceptbottleneck}. A previous human study~\cite{paleja2023interpretable} also supports the interpretability of such trees.

\subsection{Takeaways}

\edit{Our results highlight followin achievements of the proposed iTRACE framework:

\begin{itemize}
    \item Fast: By distilling VLM-generated feature extractors into a lightweight CNN, iTRACE avoids repeatedly querying the segmentation model at every timestep. This design yields several orders of magnitude speedup, making RL training practical at scale.
    \item Trainable: Both the feature extractor and the tree-based policy adapt effectively to new domains via gradient-based optimization. This enables pixels-to-action control and shows that interpretability does not come at the cost of flexibility.
    \item Interpretable: iTRACE provides transparency at every stage. The feature space is grounded in human-understandable concepts, and the learned Interpretable Control Tree ensures the resulting policy is verifiable.
\end{itemize}

In summary iTRACE delivers a pixels-to-actions system that is simultaneously fast, trainable, and interpretable.
}

\section{Limitations and future works}
\label{sec:future_work}

\polina{Our framework is limited by two factors: first, it assumes that SAM performs reasonably well across environments, which may hinder generalizability in domains with poor segmentation; second, it requires domain knowledge, which still involves human input to some extent and is left to be automated by foundation model queries or agentic exploration.}
Future work could explore RL-based backpropagation to fine-tune VLMs, allowing them to iteratively improve outputs based on agent feedback~\cite{alrashedy2024generating}. This approach has the potential to enable VLMs to identify discrepancies between the underlying physics simulation and its visual representation, leading to improved feature extraction for downstream RL tasks. 


\vspace{-1em}

\section{Conclusion}


We introduce \textbf{iTRACE}, a novel framework for achieving automated semantic interpretability in RL. iTRACE constructs interpretable feature spaces via zero-shot VLM querying and trains Interpretable Control Trees via RL to produce transparent, verifiable policies. It generalizes to unseen environments without human input and achieves performance comparable to black-box models. While all modern interpretable models, including tree-based approaches, face challenges in complex environments, iTRACE stands out among them as the first to achieve automated semantic interpretability in RL, marking a significant advancement that underscores its impact.

\vspace{-0.5em}
\bibliography{reference}

\begin{thebibliography}{10}
\providecommand{\url}[1]{#1}
\csname url@rmstyle\endcsname
\providecommand{\newblock}{\relax}
\providecommand{\bibinfo}[2]{#2}
\providecommand\BIBentrySTDinterwordspacing{\spaceskip=0pt\relax}
\providecommand\BIBentryALTinterwordstretchfactor{4}
\providecommand\BIBentryALTinterwordspacing{\spaceskip=\fontdimen2\font plus
\BIBentryALTinterwordstretchfactor\fontdimen3\font minus \fontdimen4\font\relax}
\providecommand\BIBforeignlanguage[2]{{%
\expandafter\ifx\csname l@#1\endcsname\relax
\typeout{** WARNING: IEEEtran.bst: No hyphenation pattern has been}%
\typeout{** loaded for the language `#1'. Using the pattern for}%
\typeout{** the default language instead.}%
\else
\language=\csname l@#1\endcsname
\fi
#2}}

\bibitem{Singh2022survey}
\BIBentryALTinterwordspacing
B.~Singh, R.~Kumar, and V.~P. Singh, ``Reinforcement learning in robotic applications: a comprehensive survey,'' \emph{Artificial Intelligence Review}, vol.~55, no.~2, pp. 945--990, Feb 2022. [Online]. Available: \url{https://doi.org/10.1007/s10462-021-09997-9}
\BIBentrySTDinterwordspacing

\bibitem{rudin2021interpretablemachinelearningfundamental}
\BIBentryALTinterwordspacing
C.~Rudin, C.~Chen, Z.~Chen, H.~Huang, L.~Semenova, and C.~Zhong, ``Interpretable machine learning: Fundamental principles and 10 grand challenges,'' 2021. [Online]. Available: \url{https://arxiv.org/abs/2103.11251}
\BIBentrySTDinterwordspacing

\bibitem{Glanois2024survey}
\BIBentryALTinterwordspacing
C.~Glanois, P.~Weng, M.~Zimmer, D.~Li, T.~Yang, J.~Hao, and W.~Liu, ``A survey on interpretable reinforcement learning,'' \emph{Machine Learning}, vol. 113, no.~8, pp. 5847--5890, Aug 2024. [Online]. Available: \url{https://doi.org/10.1007/s10994-024-06543-w}
\BIBentrySTDinterwordspacing

\bibitem{silva2021encoding}
A.~Silva and M.~Gombolay, ``Encoding human domain knowledge to warm start reinforcement learning,'' in \emph{Proceedings of the AAAI conference on artificial intelligence}, vol.~35, no.~6, 2021, pp. 5042--5050.

\bibitem{paleja2023interpretable}
R.~Paleja, L.~Chen, Y.~Niu, A.~Silva, Z.~Li, S.~Zhang, C.~Ritchie, S.~Choi, K.~C. Chang, H.~E. Tseng, \emph{et~al.}, ``Interpretable reinforcement learning for robotics and continuous control,'' \emph{arXiv preprint arXiv:2311.10041}, 2023.

\bibitem{delfosse2024interpretable}
Q.~Delfosse, S.~Sztwiertnia, M.~Rothermel, W.~Stammer, and K.~Kersting, ``Interpretable concept bottlenecks to align reinforcement learning agents,'' \emph{arXiv preprint arXiv:2401.05821}, 2024.

\bibitem{jiang2019neural}
Z.~Jiang and S.~Luo, ``Neural logic reinforcement learning,'' in \emph{International conference on machine learning}.\hskip 1em plus 0.5em minus 0.4em\relax PMLR, 2019, pp. 3110--3119.

\bibitem{arulkumaran2017deep}
K.~Arulkumaran, M.~P. Deisenroth, M.~Brundage, and A.~A. Bharath, ``Deep reinforcement learning: A brief survey,'' \emph{IEEE Signal Processing Magazine}, vol.~34, no.~6, pp. 26--38, 2017.

\bibitem{Liu2015feature}
\BIBentryALTinterwordspacing
D.-R. Liu, H.-L. Li, and D.~Wang, ``Feature selection and feature learning for high-dimensional batch reinforcement learning: A survey,'' \emph{International Journal of Automation and Computing}, vol.~12, no.~3, pp. 229--242, Jun 2015. [Online]. Available: \url{https://doi.org/10.1007/s11633-015-0893-y}
\BIBentrySTDinterwordspacing

\bibitem{greydanus2018visualizing}
S.~Greydanus, A.~Koul, J.~Dodge, and A.~Fern, ``Visualizing and understanding atari agents,'' in \emph{International conference on machine learning}.\hskip 1em plus 0.5em minus 0.4em\relax PMLR, 2018, pp. 1792--1801.

\bibitem{doumanoglou2023unsupervisedinterpretablebasisextraction}
\BIBentryALTinterwordspacing
A.~Doumanoglou, S.~Asteriadis, and D.~Zarpalas, ``Unsupervised interpretable basis extraction for concept-based visual explanations,'' 2023. [Online]. Available: \url{https://arxiv.org/abs/2303.10523}
\BIBentrySTDinterwordspacing

\bibitem{scabini2024comparativesurveyvisiontransformers}
\BIBentryALTinterwordspacing
L.~Scabini, A.~Sacilotti, K.~M. Zielinski, L.~C. Ribas, B.~D. Baets, and O.~M. Bruno, ``A comparative survey of vision transformers for feature extraction in texture analysis,'' 2024. [Online]. Available: \url{https://arxiv.org/abs/2406.06136}
\BIBentrySTDinterwordspacing

\bibitem{Zytek2022need}
\BIBentryALTinterwordspacing
A.~Zytek, I.~Arnaldo, D.~Liu, L.~Berti-Equille, and K.~Veeramachaneni, ``The need for interpretable features: Motivation and taxonomy,'' \emph{SIGKDD Explor. Newsl.}, vol.~24, no.~1, p. 1–13, June 2022. [Online]. Available: \url{https://doi.org/10.1145/3544903.3544905}
\BIBentrySTDinterwordspacing

\bibitem{templeton2024scaling}
A.~Templeton, T.~Conerly, J.~Marcus, J.~Lindsey, T.~Bricken, B.~Chen, A.~Pearce, C.~Citro, E.~Ameisen, A.~Jones, \emph{et~al.}, ``Scaling monosemanticity: Extracting interpretable features from claude 3 sonnet. transformer circuits thread,'' 2024.

\bibitem{yang2023foundationmodelsdecisionmaking}
\BIBentryALTinterwordspacing
S.~Yang, O.~Nachum, Y.~Du, J.~Wei, P.~Abbeel, and D.~Schuurmans, ``Foundation models for decision making: Problems, methods, and opportunities,'' 2023. [Online]. Available: \url{https://arxiv.org/abs/2303.04129}
\BIBentrySTDinterwordspacing

\bibitem{zhai2024finetuninglargevisionlanguagemodels}
\BIBentryALTinterwordspacing
Y.~Zhai, H.~Bai, Z.~Lin, J.~Pan, S.~Tong, Y.~Zhou, A.~Suhr, S.~Xie, Y.~LeCun, Y.~Ma, and S.~Levine, ``Fine-tuning large vision-language models as decision-making agents via reinforcement learning,'' 2024. [Online]. Available: \url{https://arxiv.org/abs/2405.10292}
\BIBentrySTDinterwordspacing

\bibitem{beechey2023explaining_rl_shap}
\BIBentryALTinterwordspacing
D.~Beechey, T.~M.~S. Smith, and O.~\c{S}im\c{s}ek, ``Explaining reinforcement learning with shapley values,'' in \emph{Proceedings of the 40th International Conference on Machine Learning}, ser. Proceedings of Machine Learning Research, A.~Krause, E.~Brunskill, K.~Cho, B.~Engelhardt, S.~Sabato, and J.~Scarlett, Eds., vol. 202.\hskip 1em plus 0.5em minus 0.4em\relax PMLR, 23--29 Jul 2023, pp. 2003--2014. [Online]. Available: \url{https://proceedings.mlr.press/v202/beechey23a.html}
\BIBentrySTDinterwordspacing

\bibitem{lu2025explainable}
Z.~Lu, M.~C. Gursoy, C.~K. Mohan, and P.~K. Varshney, ``Explainable ai for radar resource management: Modified lime in deep reinforcement learning,'' \emph{arXiv preprint arXiv:2506.20916}, 2025.

\bibitem{shi2022selfsupervised}
W.~Shi, G.~Huang, S.~Song, Z.~Wang, T.~Lin, and C.~Wu, ``Self-supervised discovering of interpretable features for reinforcement learning,'' \emph{IEEE Transactions on Pattern Analysis and Machine Intelligence}, vol.~44, no.~5, pp. 2712--2724, 2022.

\bibitem{wang2020attribution}
Y.~Wang, M.~Mase, and M.~Egi, ``Attribution-based salience method towards interpretable reinforcement learning.'' in \emph{AAAI Spring Symposium: Combining Machine Learning with Knowledge Engineering (1)}, 2020.

\bibitem{quinlan1987generating}
J.~R. Quinlan, ``Generating production rules from decision trees,'' in \emph{Proceedings of the 10th International Joint Conference on Artificial Intelligence - Volume 1}, ser. IJCAI'87.\hskip 1em plus 0.5em minus 0.4em\relax San Francisco, CA, USA: Morgan Kaufmann Publishers Inc., 1987, p. 304–307.

\bibitem{loh2011classification}
W.-Y. Loh, ``Classification and regression trees,'' \emph{Wiley interdisciplinary reviews: data mining and knowledge discovery}, vol.~1, no.~1, pp. 14--23, 2011.

\bibitem{hu2019optimal}
X.~Hu, C.~Rudin, and M.~Seltzer, ``Optimal sparse decision trees,'' \emph{Advances in neural information processing systems}, vol.~32, 2019.

\bibitem{grinsztajn2022tree}
L.~Grinsztajn, E.~Oyallon, and G.~Varoquaux, ``Why do tree-based models still outperform deep learning on typical tabular data?'' \emph{Advances in neural information processing systems}, vol.~35, pp. 507--520, 2022.

\bibitem{mcelfresh2023neural}
D.~McElfresh, S.~Khandagale, J.~Valverde, V.~Prasad~C, G.~Ramakrishnan, M.~Goldblum, and C.~White, ``When do neural nets outperform boosted trees on tabular data?'' \emph{Advances in Neural Information Processing Systems}, vol.~36, pp. 76\,336--76\,369, 2023.

\bibitem{zhou2019deep}
Z.-H. Zhou and J.~Feng, ``Deep forest,'' \emph{National science review}, vol.~6, no.~1, pp. 74--86, 2019.

\bibitem{silva_optimization_2020}
\BIBentryALTinterwordspacing
A.~Silva, T.~Killian, I.~D.~J. Rodriguez, S.-H. Son, and M.~Gombolay, ``Optimization {Methods} for {Interpretable} {Differentiable} {Decision} {Trees} in {Reinforcement} {Learning},'' June 2020, arXiv:1903.09338 [cs, stat]. [Online]. Available: \url{http://arxiv.org/abs/1903.09338}
\BIBentrySTDinterwordspacing

\bibitem{paleja_learning_2023}
\BIBentryALTinterwordspacing
R.~Paleja, Y.~Niu, A.~Silva, C.~Ritchie, S.~Choi, and M.~Gombolay, ``Learning {Interpretable}, {High}-{Performing} {Policies} for {Autonomous} {Driving},'' July 2023, arXiv:2202.02352 [cs]. [Online]. Available: \url{http://arxiv.org/abs/2202.02352}
\BIBentrySTDinterwordspacing

\bibitem{pace2022poetree}
A.~Pace, A.~J. Chan, and M.~van~der Schaar, ``Poetree: Interpretable policy learning with adaptive decision trees,'' \emph{arXiv preprint arXiv:2203.08057}, 2022.

\bibitem{wu2024readreaprewardslearning}
\BIBentryALTinterwordspacing
Y.~Wu, Y.~Fan, P.~P. Liang, A.~Azaria, Y.~Li, and T.~M. Mitchell, ``Read and reap the rewards: Learning to play atari with the help of instruction manuals,'' 2024. [Online]. Available: \url{https://arxiv.org/abs/2302.04449}
\BIBentrySTDinterwordspacing

\bibitem{liu2024surveyhallucinationlargevisionlanguage}
\BIBentryALTinterwordspacing
H.~Liu, W.~Xue, Y.~Chen, D.~Chen, X.~Zhao, K.~Wang, L.~Hou, R.~Li, and W.~Peng, ``A survey on hallucination in large vision-language models,'' 2024. [Online]. Available: \url{https://arxiv.org/abs/2402.00253}
\BIBentrySTDinterwordspacing

\bibitem{peng2024preference}
\BIBentryALTinterwordspacing
A.~Peng, A.~Bobu, B.~Z. Li, T.~R. Sumers, I.~Sucholutsky, N.~Kumar, T.~L. Griffiths, and J.~A. Shah, ``Preference-conditioned language-guided abstraction,'' 2024. [Online]. Available: \url{https://arxiv.org/abs/2402.03081}
\BIBentrySTDinterwordspacing

\bibitem{yang2023finegrained}
\BIBentryALTinterwordspacing
L.~Yang, Y.~Wang, X.~Li, X.~Wang, and J.~Yang, ``Fine-grained visual prompting,'' 2023. [Online]. Available: \url{https://arxiv.org/abs/2306.04356}
\BIBentrySTDinterwordspacing

\bibitem{Kirillov_2023_Segment_Anything}
A.~Kirillov, E.~Mintun, N.~Ravi, H.~Mao, C.~Rolland, L.~Gustafson, T.~Xiao, S.~Whitehead, A.~C. Berg, W.-Y. Lo, P.~Dollar, and R.~Girshick, ``Segment anything,'' in \emph{Proceedings of the IEEE/CVF International Conference on Computer Vision (ICCV)}, October 2023, pp. 4015--4026.

\bibitem{li2025mihbench}
J.~Li, M.~Wu, Z.~Jin, H.~Chen, J.~Ji, X.~Sun, L.~Cao, and R.~Ji, ``Mihbench: Benchmarking and mitigating multi-image hallucinations in multimodal large language models,'' \emph{arXiv preprint arXiv:2508.00726}, 2025.

\bibitem{jang2016categorical}
E.~Jang, S.~Gu, and B.~Poole, ``Categorical reparameterization with gumbel-softmax,'' \emph{arXiv preprint arXiv:1611.01144}, 2016.

\bibitem{bengio2013estimating}
Y.~Bengio, N.~L{\'e}onard, and A.~Courville, ``Estimating or propagating gradients through stochastic neurons for conditional computation,'' \emph{arXiv preprint arXiv:1308.3432}, 2013.

\bibitem{carlini2023quantifyingmemorizationneurallanguage}
\BIBentryALTinterwordspacing
N.~Carlini, D.~Ippolito, M.~Jagielski, K.~Lee, F.~Tramer, and C.~Zhang, ``Quantifying memorization across neural language models,'' 2023. [Online]. Available: \url{https://arxiv.org/abs/2202.07646}
\BIBentrySTDinterwordspacing

\bibitem{sahoo2018learning}
S.~Sahoo, C.~Lampert, and G.~Martius, ``Learning equations for extrapolation and control,'' in \emph{International Conference on Machine Learning}.\hskip 1em plus 0.5em minus 0.4em\relax Pmlr, 2018, pp. 4442--4450.

\bibitem{schulman2017proximal}
J.~Schulman, F.~Wolski, P.~Dhariwal, A.~Radford, and O.~Klimov, ``Proximal policy optimization algorithms,'' \emph{arXiv preprint arXiv:1707.06347}, 2017.

\bibitem{pang-wei2020conceptbottleneck}
P.~W. Koh, T.~Nguyen, Y.~S. Tang, S.~Mussmann, E.~Pierson, B.~Kim, and P.~Liang, ``Concept bottleneck models,'' in \emph{Proceedings of the 37th International Conference on Machine Learning}, ser. Proceedings of Machine Learning Research, H.~D. III and A.~Singh, Eds., vol. 119.\hskip 1em plus 0.5em minus 0.4em\relax PMLR, 13--18 Jul 2020, pp. 5338--5348.

\bibitem{alrashedy2024generating}
K.~Alrashedy, P.~Tambwekar, Z.~Zaidi, M.~Langwasser, W.~Xu, and M.~Gombolay, ``Generating cad code with vision-language models for 3d designs,'' \emph{arXiv preprint arXiv:2410.05340}, 2024.

\end{thebibliography}
\bibliographystyle{IEEEtran}


\end{document}